\pdfoutput=1

\documentclass[11pt]{article}

\usepackage{authblk}

\usepackage{ACL2023}

\usepackage{times}
\usepackage{latexsym}

\usepackage[T1]{fontenc}

\usepackage[utf8]{inputenc}

\usepackage{microtype}

\usepackage{inconsolata}

\usepackage{soul}
\usepackage{url}
\usepackage{breakurl}

\usepackage[utf8]{inputenc}
\usepackage[font=footnotesize]{caption}
\usepackage{graphicx}
\usepackage{amsmath,amsfonts} 
\usepackage{booktabs}
\usepackage{subcaption}
\usepackage{mathtools}
\usepackage{makecell}

\usepackage{tabu}
\usepackage{stfloats}

\usepackage{enumitem}
\usepackage{siunitx}
\usepackage{multirow}
\usepackage{algorithm}
\usepackage{setspace}
\usepackage{algcompatible}
\usepackage[noend]{algpseudocode}
\makeatletter
\def\BState{\State\hskip-\ALG@thistlm}
\makeatother

\algdef{SE}[SUBALG]{Indent}{EndIndent}{}{\algorithmicend\ }%
\algtext*{Indent}
\algtext*{EndIndent}

\usepackage{xcolor}
\newcommand{\parm}{\mathord{\color{black!33}\bullet}}

\newcommand\METHOD{RST~} 
\newcommand\ADRDT{ADR~}
\newcommand\PRODUCTDT{Product~}
\newcommand\AMAZONDT{Amazon~}
\newcommand\YELPDT{Yelp~}
\newcommand\AGNEWSDT{AG-News~}

\title{Neural Networks Against (and For) Self-Training: Classification with Small Labeled and Large Unlabeled Sets}

\author[1,2]{Payam Karisani}
\affil[1]{University of Illinois at Urbana-Champaign}
\affil[2]{Emory University}
\affil[ ]{ {\fontfamily{pcr}\selectfont payam.karisani@emory.edu} }

\begin{document}
\maketitle
\begin{abstract}

We propose a semi-supervised text classifier based on self-training using one positive and one negative property of neural networks. One of the weaknesses of self-training is the semantic drift problem, where noisy pseudo-labels accumulate over iterations and consequently the error rate soars. In order to tackle this challenge, we reshape the role of pseudo-labels and create a hierarchical order of information. In addition, a crucial step in self-training is to use the classifier confidence prediction to select the best candidate pseudo-labels. This step cannot be efficiently done by neural networks, because it is known that their output is poorly calibrated. To overcome this challenge, we propose a hybrid metric to replace the plain confidence measurement. Our metric takes into account the prediction uncertainty via a subsampling technique. We evaluate our model in a set of five standard benchmarks, and show that it significantly outperforms a set of ten diverse baseline models. Furthermore, we show that the improvement achieved by our model is additive to language model pretraining, which is a widely used technique for using unlabeled documents. Our code is available at \url{https://github.com/p-karisani/RST}.

\end{abstract}

\section{Introduction} \label{sec:intro}

Text classification has achieved tremendous success in the past decade thanks to the advancement in deep neural networks. Even though the introduction of contextual word embeddings and language model pretraining \cite{elmo,bert} has greatly reduced the reliance on large manually annotated datasets, the current over-parametrized models are still prone to over-fitting. 
To further reduce this reliance one can use unlabeled data \cite{semiupervised-nlp-book,semi-supervised-book}. 
In this article, we use the properties of neural networks and develop a self-training model, termed Robust Self-Training (\METHOD\!), for low-data regime text classification. Self-training (also known as pseudo-labeling) is iterative \cite{self-train-0,neural-pseudo-label}, and in each iteration unlabeled documents are automatically annotated and augmented with labeled documents.

Previous studies \cite{nell,semantic-drift-solution} report that self-training suffers from the semantic drift problem. That is, as the iterations are carried on, spurious pseudo-labels are generated and added to the labeled documents. This eventually distorts the class boundaries and drifts the original class centroids. To address this problem, inspired by the catastrophic forgetting phenomenon in neural networks \cite{catastrophic-1}, we propose a novel procedure to reshape the role of pseudo-labels in the algorithm. We also aim to overcome a weakness of neural networks in this algorithm. Self-training relies on prediction confidence to select the best candidate documents. In this framework, the classifier output is interpreted as prediction confidence \cite{neural-self-train-baselines}. Self-training performance deteriorates in the settings that the used classifier is unable to accurately estimate the prediction confidence \cite{UPS}. Neural networks suffer from such a problem, because their outputs are mis-calibrated \cite{neural-conf-0}. To address this problem, we propose a novel metric to replace the plain confidence measurement. Our metric takes into account the prediction uncertainty via a subsampling algorithm.


We use a set of five standard benchmarks to evaluate our model. The selected datasets cover a wide spectrum of documents, ranging from formal news documents to highly informal social media documents. We also compare our model with a set of ten methods, including approaches that use variants of self-training, use multiple classifiers, use multi-view learning, and use various uncertainty metrics. The experiments signify to the superiority of our model. Additionally, we analyze our model and demonstrate that the improvement achieved by our model is additive to the performance of domain specific language model pretraining.

The contributions of our work are as follows: \textbf{1)} We mitigate the semantic drift problem in self-training by reshaping the role of pseudo-labeled documents and creating a hierarchical order of information. \textbf{2)} We enhance the pseudo-label selection in self-training by proposing a novel selection metric to replace the plain confidence measurement. Our metric is particularly advantageous when neural networks are used as the underlying classifier, because these classifiers are overconfident in their predictions \cite{neural-conf-0}. \textbf{3)} Through an extensive set of experiments with five datasets and ten baselines we show that our model is highly resistant to noisy pseudo-labels, yields an additive improvement to domain specific language model pretraining, and outperforms the state of the art.



\section{Related Work} \label{sec:rel-work}

\noindent\textbf{Neural networks in self-training.} Self-training or pseudo-labeling \cite{self-train-0,neural-pseudo-label} is a semi-supervised learning algorithm. Previous studies investigate various aspects of this algorithm and aim for filling the niches. For instance, \citet{deep-pseudo-label} integrate MixUp \cite{mixup} with the oversampling of labeled documents, \citet{self-train-queue} proposes a new document sampling strategy, \citet{noise-self-train} and \citet{self-train-two-staged} report that adding noise to pseudo-labels and the hidden layers of a network enhances the model performance--the latter for the sequence generation task, and \citet{self-train-vs-self-pretrain} contrast self-training and pretraining and conclude that under certain conditions the former outperforms the latter. \citet{co-decomp} propose a multi-view self-training model to incorporate domain-knowledge, \citet{meta-pseudo} propose a feedback loop across self-training iterations, \citet{ruled-self-train} propose a model to incorporate weakly supervised domain specific rules, \citet{nli-self-train} report that pre-training a model with an auxiliary NLI task enhances self-training, and \citet{bdd} reduces the variance of pseudo-labels within each class using an angular loss.

As opposed to our research, none of these studies propose a model to maintain a balance between the set of pseudo-labels and the set of manual labels. Additionally, they don't analyze the deterioration of performance during the self-training iterations, and consequently have no defense against this fundamental weakness.

\noindent\textbf{Uncertainty measurement in NLP.} Confidence in model prediction, is the amount of trust in the predicted class label compared to the other class labels \cite{neural-conf-0}. Uncertainty in model prediction, is the amount of trust in the entire prediction regardless of the predicted label \cite{uncertainty}. The research on the efficacy of uncertainty in semi-supervised learning is scarce. \citet{UST} propose to filter out uncertain predictions before the candidate selection step, \citet{UPS} apply a set of thresholds to filter out uncertain and unconfident predictions, and \citet{HAU} experiment with various uncertainty metrics and report that uncovering the Heteroscedastic uncertainty (the intrinsic data uncertainty) \cite{uncertainty} is the best strategy on average.

As opposed to our work, none of these studies propose an integrated metric for selecting pseudo-labels. Additionally, after selecting the pseudo-labels, they don't propose any strategy to restrain the noisy labels from polluting the labeled set.

\noindent\textbf{Ensemble and multi-view models.} There exist models that use multiple classifiers, examples include variants of Tri-training \cite{tri-train-d,neural-self-train-baselines}, variants of co-training \cite{co-train,co-regular,reinforced-co-train,co-decomp}, and other ad hoc ensemble models \cite{co-forest,cotrain-bag,mutual-learn}.

As opposed to our work, these models rely only on the confidence of classifiers. No coherent uncertainty interpretation has been proposed for them. Additionally, they use ensembling in the prediction stage, whereas, we employ only one classifier for this purpose, which is more resource efficient.

\noindent\textbf{Semantic drift in self-training.} Semantic drift in self-training \cite{semantic-drift,nell} occurs when spurious pseudo-labels accumulate over time and distort the distribution of labeled documents. In the context of neural networks, the research in this area is sparse. One approach is to avoid pseudo-labels altogether, and use unlabeled documents differently \cite{var-pretrain,UDA,mixtext,do-pretrain}. Nonetheless, these alternative methods don't necessarily compete with self-training and can co-exist with it inside a framework. To address semantic drift directly, existing approaches mainly aim for explicitly adjusting pseudo-labels. \citet{bdd} use an angular loss function to project pseudo-labels. \citet{self-pretraining} assume pseudo-labels evolve in a stochastic process and normalize their values. In terms of the role of pseudo-labels in self-training, our algorithm can be taken as the generalized form of the algorithm proposed by \citet{self-pretraining}.

\noindent\textbf{Connections to consistency regularization.} There are two distinctions between our model and consistency training methods \cite{contarstive-learn,UDA}, one in the methodology and another in the objective. In consistency-based regularization methods, data points are manipulated and new data points are generated. As opposed to these methods we don’t manipulate data, instead we revise the training steps. Additionally, in consistency based regularization methods the goal is to create a smooth loss surface, so that the class boundaries are easier to adjust and expand to unlabeled data. Our objective is different, we aim to address the model overconfidence, which is why we don’t use this step during the training of the classifier (consistency regularization is done during the training), we use it only during the candidate selection. This means that our method doesn't compete with consistency training, and can co-exist with it in a single framework.

\section{Proposed Method} \label{sec:method}

In a typical self-training model \cite{self-train}, there is a set $L$ of labeled data, and a set $U$ of unlabeled data. A predictive model is trained on $L$ and is used to probabilistically label $U$. Then, given a hyper-parameter $\theta$, as the minimum confidence threshold, the confidently labeled documents in $U$ and their associated \textit{pseudo-labels} are selected and added to $L$. This procedure is iterative. In this framework, there is no constraint on the choice of the underlying model, except that it is required to assign a confidence score to each pseudo-label. 

There are two challenges to face in this setting:~1)~Self-training suffers from the semantic drift problem \cite{semantic-drift,nell}. That is, as we increase the number of iterations, the error rate accumulates and the class boundaries are distorted. 2)~Neural networks are overconfident in their predictions \cite{neural-conf-0,neural-conf-1}, and as we discuss in Section~\ref{sub-sec:subsample}, this shortcoming deteriorates the quality of the pseudo-labels in each iteration.

To address these challenges, we present Robust Self-Training (\METHOD\!). Algorithm \ref{alg:summary} provides an overview of \METHOD, with two classifiers, in Structured English. Lines \ref{alg-line:self-begins} to \ref{alg-line:self-ends} demonstrate one iteration of the algorithm, which is repeated till the set of unlabeled documents $U$ is exhausted. The iteration begins by initializing the classifiers $C_{1}$ and $C_{2}$. Then, it continues by sampling from the set of pseudo-labels $S$ and distilling it \cite{distill} into the classifier $C_{1}$.\footnote{In the first iteration the set $S$ is empty, therefore, no distillation is done.} Then, another sample from the set of labeled documents~$L$ is taken to further train $C_{1}$ using Equation \ref{eq:loss} (see Section \ref{sub-sec:pretrain}). These steps are re-taken to train the second classifier $C_{2}$. Finally, $C_{1}$ and $C_{2}$ are used in Equation \ref{eq:score} (see Section~\ref{sub-sec:subsample}) to label and score the documents in the set $U$. The top documents are removed from $U$ and are added to the set $S$. Since we have multiple classifiers labeling each document (in this case two classifiers), we store the average of the outputs in $S$. On Line~\ref{alg-line:pretrain}, the entire set~$S$ is used to pretrain the final classifier~$C$, and on Line~\ref{alg-line:finetune}, the set~$L$ is used to finetune~$C$ using Equation~\ref{eq:loss}. In the following two sections we discuss how our algorithm can address the two aforementioned challenges.

\begin{algorithm}
\footnotesize
\caption{Overview of \METHOD}\label{alg:summary}
\begin{algorithmic}[1]
\algrenewcommand\algorithmicindent{7pt}
\Procedure{RST}{}
    \State \textbf{Given:}
    \Indent
        \State $L: \text{Set of labeled documents}$ 
        \State $U: \text{Set of unlabeled documents}$
    \EndIndent
    \State \textbf{Return:}
    \Indent
        \State $\text{Trained classifier on L and U}$
    \EndIndent
    \State \textbf{Execute:}
    \Indent
        \State Set $K$ to 100 // $\text{hyper-parameter (step size)}$
        \State Set $R$ to 70 // $\text{hyper-parameter (sample ratio)}$
        \State Set $S$ to EMPTY // $\text{the set of pseudo-labels}$
        \While{$U$ is not EMPTY}
            \State Initialize \parbox[t]{.7\linewidth}{the classifiers $C_{1}$ and $C_{2}$}\label{alg-line:self-begins}
            \State Sample \parbox[t]{.7\linewidth}{$R\%$ of $S$, order the data as described in Section \ref{sub-sec:pretrain}, and use it to train $C_{1}$}
            \State Sample \parbox[t]{.7\linewidth}{$R\%$ of $L$ and use in Equation \ref{eq:loss} for $C_{1}$}
            \State Sample \parbox[t]{.7\linewidth}{$R\%$ of $S$, order the data as described in Section \ref{sub-sec:pretrain}, and use it to train $C_{2}$}
            \State Sample \parbox[t]{.7\linewidth}{$R\%$ of $L$ and use in Equation \ref{eq:loss} for $C_{2}$}
            \State Use $C_{1}$ \parbox[t]{.7\linewidth}{and $C_{2}$ to label $U$ and then score the documents using Equation \ref{eq:score}}
            \State Remove \parbox[t]{.7\linewidth}{the top $K$ documents from $U$ and add them to $S$}
            \EndWhile\label{alg-line:self-ends}
        \State Order \parbox[t]{.7\linewidth}{the set $S$ as described in Section \ref{sub-sec:pretrain}, and use it to train the classifier $C$\label{alg-line:pretrain}}
        \State Use the set $L$ in Equation \ref{eq:loss} to further train $C$\label{alg-line:finetune}
        \State \textbf{Return} $C$
    \EndIndent
\EndProcedure
\end{algorithmic}
\end{algorithm}

\subsection{Overcoming Semantic Drift} \label{sub-sec:pretrain}
An inherent pitfall of the self-training algorithm is the semantic drift problem \cite{semantic-drift,nell,semantic-drift-solution}, where adding new pseudo-labels ultimately impacts the properties of the classes in the set of labeled documents. 
To mitigate this problem, one solution is to order the training data based on the deemed noise in the labels.\footnote{One can also reduce the importance of the pseudo-labels, which we use as a baseline.} Thus, we seek to re-design self-training to undergo such a modification.

Catastrophic forgetting is a problem in neural networks \cite{catastrophic-1, catastrophic-2}. This problem arises in the continual learning settings, when a series of tasks are sequentially used to train a neural network. It stems from the fact that the weights of the neural network update to serve the objective of the current task, therefore, the information about the current task replaces the information about the previous tasks. We use this property of neural networks to construct a natural hierarchical order of information. Because the pseudo-labels in each iteration are obtained by the model of the previous iteration, it is reasonable to assume that they are noisier than the pseudo-labels in the previous iterations. Based on this argument, we propose to order the pseudo-labels according to the reverse iteration number, and then, use them to train the network of the current iteration. Because it is assumed that the labeled data is noiseless, this set is used at the end of the training to finetune the network. One can assume that the pseudo-labels in this algorithm are used to initialize the network, and the labeled data is used to finetune the network.


To be able to initialize and train the network in each iteration, we store the iteration number that each pseudo-label was added to the pool. We call the set of pseudo-labels the set $S$, and the set of initial labeled documents the set $L$. At the beginning of each iteration, we order the pseudo-labels in $S$ and use them to train the network, i.e., Task~1. We store--and use--the last layer logits of the network in classifying the documents in $S$ to be used with a high temperature for initialization. Thus, we essentially distill the knowledge of the previous iterations into the network \cite{distill}. Additionally, because randomness in creating the batches is an essential ingredient of stochastic gradient descent, while training the network by the pseudo-labels of each iteration, we randomly select a percentage of pseudo-labels from other iterations. Finally, we use the documents in $L$ and minimize the following objective function to further train the network, i.e., Task~2:
\begin{equation} \label{eq:loss}
\setlength{\jot}{0pt}
\setlength{\abovedisplayskip}{6pt}
\setlength{\belowdisplayskip}{6pt}
\medmuskip=0mu
\thinmuskip=0mu
\thickmuskip=0mu
\nulldelimiterspace=0pt
\scriptspace=0pt
\begin{split}
\mathcal{L}= &(1-\lambda)(-\sum_{i=1}^{N}[y_{i}\log a_{i}+(1-y_{i})\log (1-a_{i})]) + \\ 
&\lambda(-\sum_{i=1}^{N}[q_{i}\log a'_{i}+(1-q_{i})\log (1-a'_{i})]),
\end{split}
\end{equation}
where $N$ is the number of the documents in the set $L$, $y_{i}$ is the binary ground truth label of the document $d_{i}$, $a_{i}$ is the output of the network after the softmax layer, $a'_{i}$ is the output of the network after the softmax layer with a high temperature \cite{distill}, and $q_{i}$ is the output of the network with the same temperature as the previous case right before the current task begins. Note that $a'_{i}$ and $q_{i}$ are different, the former refers to the current output of the network, while the weights are still being updated to meet the objective of the current task, and the latter refers to the output of the network after it was trained by the documents in the set $S$ and before it was trained by the documents in the set $L$. $\lambda$ is a penalty term ($0 \leq \lambda \leq 1$).

The first term in the loss function is the regular cross entropy between the ground truth labels and the output class distributions. The second term is the cross entropy between the current output class distributions and the class distributions that were obtained after training the network by the pseudo-labels. Intuitively, the goal of the second term is to prevent the first term from fully erasing the knowledge already stored in the network, i.e., the knowledge obtained from the pseudo-labels.


One advantage of employing pseudo-labels to initialize the network, as we described in this section, is that if during the self-training iterations due to the growing size of the set $S$ the newly added pseudo-labels become highly noisy, the first term in the objective function yields stronger gradients and will automatically dampen the effect of these examples. In fact, we show that given this mechanism, there is no need to validate the number of self-training iterations anymore, and one can label and use the entire set $U$. Whereas, doing so in the regular self-training causes semantic drift.


\subsection{Addressing Overconfidence} \label{sub-sec:subsample}
The performance in each self-training iteration heavily depends on the quality of the pseudo-labels added to the training data in the previous iterations. Neural networks are prone to assigning high posterior probabilities even to the out of distribution data points \cite{neural-conf-1}. This means, with a high probability, the mislabeled documents can be confidently assigned to the opposite class and can be selected as the best candidate documents. The effect of these spurious labels can accumulate over iterations and eventually deteriorate the performance.\footnote{Throttling is used to reduce this effect \cite{semiupervised-nlp-book}, which we use in the experiments. Our solution specifically reduces the noise introduced by an overconfident model.}

To address this issue, in this section we propose a novel selection criterion to replace the plain confidence metric. Our criterion takes into account the uncertainty in the classifier output. The core idea of our algorithm is to determine whether the output class distributions of a candidate document under multiple different subsamples of the set $L$ are consistent.\footnote{In terms of runtime and memory consumption our model is comparable to existing semi-supervised learning models. In fact, three of our baselines consist of model ensembling (they use two or more classifiers), which is a common practice in semi-supervised learning.} A small divergence--while having distinctly different training sets--indicates that there are strong similarities between the candidate document and the set $L$. A high confidence that occurs due to the poor calibration of neural network outputs, and not because of the qualities of data, is less likely to re-occur under multiple sample sets.

To implement this idea, we note that the selection criterion must be proportional to model confidence and disproportional to output uncertainty. Below we propose a metric that follows our desired criteria:


\begin{equation} \label{eq:score}
\setlength{\jot}{0pt}
\setlength{\abovedisplayskip}{6pt}
\setlength{\belowdisplayskip}{6pt}
\medmuskip=0mu
\thinmuskip=0mu
\thickmuskip=0mu
\nulldelimiterspace=0pt
\scriptspace=0pt
\begin{split}
Score(d)=\frac{\prod_{i=1}^{m}(1 - \hat{H}(P_{a_{i}})) + \alpha }{GJS(P_{a_{1}}, \dots ,P_{a_{m}}) + \alpha},
\end{split}
\end{equation}

where $d$ is the candidate document; $P_{a_{i}}$ is the output distribution of the classifier $C_{i}$ trained on the \textit{i-th} subsample; $\hat{H}(a_{i})$ is the normalized entropy of the class distribution; $GJS$ is the generalized Jensen-Shannon distance between the class distributions $P_{a_{1}}, \dots, P_{a_{m}}$; $m$ is the number of subsamples---in Algorithm \ref{alg:summary} $m$ equals 2---; and $\alpha$ is a smoothing factor---we set it to \num{1e-4} in all the experiments. Depending on the value of $\alpha$, the equation results in $Score(d)\in(0,+\infty)$.

The normalized entropy \cite{n-entropy} of a random variable is the entropy of the random variable divided by its maximum entropy:
\begin{equation} \nonumber
\setlength{\jot}{0pt}
\setlength{\abovedisplayskip}{6pt}
\setlength{\belowdisplayskip}{6pt}
\medmuskip=0mu
\thinmuskip=0mu
\thickmuskip=0mu
\nulldelimiterspace=0pt
\scriptspace=0pt
\begin{split}
\hat{H}(X) = -\sum_{}^{n} p(X) \frac{\log~p(X)}{\log~n},
\end{split}
\end{equation}
where $n$ is the number of classes. We use the normalized variant instead of the regular Shannon entropy to scale the quantity between 0 and 1. The generalized Jensen-Shannon distance \cite{gjs} measures the diversity between a set of distributions, and is calculated as follows:

\begin{equation} \nonumber
\setlength{\jot}{0pt}
\setlength{\abovedisplayskip}{6pt}
\setlength{\belowdisplayskip}{6pt}
\medmuskip=0mu
\thinmuskip=0mu
\thickmuskip=0mu
\nulldelimiterspace=0pt
\scriptspace=0pt
\begin{split}
GJS(P_{a_{1}}, \dots ,P_{a_{m}})=H(\overline{P})-\frac{1}{m} \sum_{i=1}^{m} H(P_{a_{i}}),
\end{split}
\end{equation}

where $H(\parm)$ is the Shannon entropy, and $\overline{P}$ is the mean of the distributions. The mean is calculated as follows:

\begin{equation} \nonumber
\setlength{\jot}{0pt}
\setlength{\abovedisplayskip}{6pt}
\setlength{\belowdisplayskip}{6pt}
\medmuskip=0mu
\thinmuskip=0mu
\thickmuskip=0mu
\nulldelimiterspace=0pt
\scriptspace=0pt
\begin{split}
\overline{P}=\frac{1}{m} \sum_{i=1}^{m} P_{a_{i}}.
\end{split}
\end{equation}

The numerator in Equation \ref{eq:score} represents the confidence of the classifiers. Higher confidence in the classification yields lower entropy in the class predictions, and hence, results in a higher score. The denominator in Equation \ref{eq:score} represents the output uncertainty. Using Equation \ref{eq:score} we can score the documents in the set $U$, and select the top documents and their associated pseudo-labels to be added to the set $L$--we assume all classifiers agree on the labels of the top candidate documents.\footnote{We did not observe an example that violates this assumption in the experiments. Nonetheless, such an example can be taken as noise and can be discarded.}

So far we discussed binary classification problems. Extending our method to multi-class tasks is trivial. To do so, we only need to replace the binomial cross entropy in Equation \ref{eq:loss} with a multinomial cross entropy. Note that Equation \ref{eq:score} remains intact, because it is agnostic to the number of classes.

\subsection{Computational Complexity} \label{sub-sec:resource}

During the experiments, we observed that even with two subsamples our model outperforms existing baselines. Therefore, we used only two classifiers in all the experiments. In terms of implementation, our model has two variants: a sequential variation and a parallel variation. In the sequential setting, the classifier $C_{1}$ is trained on the sets $S$ and $L$, and then it is used to label the set $U$. The pseudo-labels are stored and the classifier is removed from the memory. This process is repeated for the classifier $C_{2}$ to obtain the second set of pseudo-labels. The two sets of pseudo-labels are processed using Equation 1, and the sets $S$ and $U$ are updated. In this setting, the memory footprint is identical to that of the regular self-training and the run-time is $2\times$ slower; because each iteration involves training both networks. In the parallel setting, both classifiers $C_{1}$ and $C_{2}$ can be trained at the same time to obtain the sets of pseudo-labels. In this case, our model has $2\times$ more parameters, because both networks should be stored in memory. Since in the parallel case the two networks do not communicate, the run-time is significantly shorter than the sequential case--it is easily comparable to that of the regular self-training.

\section{Experimental Setup} \label{sec:setup}

In the current and in the next sections we describe our experimental setup and our results. 

\subsection{Datasets}\label{subsec:datasets}
We evaluate our model in the sentiment analysis task, in the news classification task, in detecting the reports of  medical drug side-effects (the ADR task), and in detecting the reports of product consumption.

In the sentiment analysis task, we use the Amazon dataset \cite{amazon-dt} and the Yelp dataset \cite{sentiment-dataset}. In the news classification task, we use the AG-News dataset \cite{agnews-dt} which is a multi-class classification task with four classes. In the ADR task, we use the dataset introduced by \citet{SMM4H-2019} prepared for an ACL 2019 Shared Task. In the product consumption task, we use the dataset introduced by \citet{vaccine-dt}. We specifically use a diverse set of datasets in the experiments to comprehensively evaluate our model. The datasets cover short and long documents. They also cover balanced, imbalanced, and extremely imbalanced tasks. They contain a multi-class task. They also contain social media classification tasks, which reportedly suffer from noisy content \cite{our-corona,al-social}.

The \AMAZONDT dataset is accompanied by a set of unlabeled documents. In \YELPDT and \AGNEWSDT datasets (for each one separately) we take a set of 10K unused training documents as unlabeled data. For \ADRDT and \PRODUCTDT datasets (for each one separately) we used the Twitter API and collected 10K in-domain documents\footnote{We used a set of related keywords to collect the documents. Depending on the subject, collecting this number of documents may take between a few days to a few weeks. It took us about 10 days to collect 10K dissimilar related documents.} to be used by the models as unlabeled data. 

\subsection{Baselines}\label{subsec:baselines}


We compare our model with a set of ten diverse models. 

\noindent\textbf{\textit{Baseline (2019).}} We include the pretrained BERT model (base version) followed by one layer fully connected network, and a softmax layer \cite{bert,bert-impl}. We follow the settings suggested in the reference to set-up the model. This baseline is finetuned on the training set and evaluated on the test set.

\noindent\textbf{\textit{Self-train (1995, 2018).}} We include the neural self-training model \cite{self-train,neural-self-train-baselines}. Based on the confidence of the classifier the top candidate pseudo-labels are selected and added to the labeled data--see the next section for the details. 
We use one instance of \textit{Baseline} as the classifier in this model.

\noindent\textbf{\textit{Tri-train+ (2010, 2018).}} We include the model introduced by \citet{tri-train-d} called tri-training with disagreement. This model is the enhanced variant of tri-training model \cite{tri-train}, and was shown to be more efficient \cite{neural-self-train-baselines}. We use three instantiations of \textit{Baseline} with different initializations in this model.

\noindent\textbf{\textit{Mutual-learn (2018).}} We include the model introduced by \citet{mutual-learn}. This model is based on the idea of raising the entropy of neural predictions to improve generalization \cite{entropy-inc}. We use two instantiations of \textit{Baseline} with different initializations in this model.

\noindent\textbf{\textit{Spaced-rep (2019).}} We include the model introduced by \citet{self-train-queue}. This model is based on the Leitner learning system. In each iteration it selects the easiest and most informative documents.

\noindent\textbf{\textit{Co-Decomp (2020).}} We include the model introduced by \citet{co-decomp}. In this model, which is a multi-view semi-supervised method, the task is decomposed into a set of sub-tasks, and then, their results are aggregated. We use two instantiations of \textit{Baseline} in this model.

\noindent\textbf{\textit{HAU (2021).}} \citet{HAU} experiment with various uncertainty and confidence measurement methods in two tasks, and report that on average Aleatoric Heteroscedastic Uncertainty metric outperforms other measurement methods. We include this method in our experiments. 

\noindent\textbf{\textit{UPS (2021).}} We include the model proposed by \citet{UPS}. This model uses a gating mechanism using thresholds to filter out uncertain and unconfident pseudo-labels. Then uses the regular cross entropy for the most confident data points, and another loss called negative cross entropy for the least confident data points. 

\noindent\textbf{\textit{BDD (2021).}} We include the model introduced by \citet{bdd}. This model uses an angular loss function to reduce the variance of label angels by transforming the values of pseudo-labels. Their hypothesis is that reducing the variance of model predictions should enhance model performance.


\noindent\textbf{\textit{Sel-Reg (2022).}} We include the method by \citet{sel-reg}. They propose a regularizer to reduce the confirmation bias in successive pseudo-labeling iterations. Their core idea is to diversify the selection of pseudo-labels using an entropy-based loss term. 
\subsection{Experimental Details}\label{subsec:exp-details}
In all the models we use pretrained BERT (the base variant) as the underlying classifier. This setting, which is realistic, makes any improvement over the naive baseline very difficult, because BERT already performs well with small labeled data \cite{bert}. On the other hand, because all the models have an identical pretrained network their comparison is completely fair.

All the models employ throttling \cite{semiupervised-nlp-book} with confidence thresholding--minimum of 0.9 as the cutoff. We also use a model similar to linear growth sampling \cite{sample-linear} for augmenting the labeled data with unlabeled data, i.e., in each iteration, we sample at most 10\% of the current set of labeled data. 
We use the optimizer suggested by \citet{bert} with the batch size of 32--Adam with a linear scheduler.

Augmenting the entire set of unlabeled data with labeled data causes semantic drift in \textit{self-training}. \citet{co-decomp} show that \textit{Co-Decomp} suffers from the same problem. Thus, we treated the number of pseudo-labels as the hyper-parameter in these models and in each experiment used 20\% of the training set as the validation set to find the best value. We tuned all of the models for the F1 measure. We found that the optimal values depend on the task and the training sets. \textit{Tri-training+} has an internal stopping criterion, and \textit{Mutual-learn} uses the entire set of unlabeled data to regulate the confidences of the two classifiers. \textit{Spaced-rep} and \textit{BDD} rely on a validation set for candidate selection. Thus, we allocated 20\% of the labeled set for this purpose. 
The rest of the settings are identical to what is suggested by \citet{self-train-queue} and \citet{bdd}. 

There are four hyper-parameters in our model: the value of softmax temperature in the distillation processes, the ratio of sampling, the value of $\lambda$ in the objective function (Equation \ref{eq:loss}), and the number of classifiers. We set the values of the temperature and the sample size to 2 and 70\% respectively across all the experiments. We tuned the value of $\lambda$ in \PRODUCTDT training set, and fixed it across all the experiments--$\lambda \in\{0.1, 0.3, 0.5, 0.7, 0.9\}$. The optimal value of $\lambda$ is 0.3, which assigns a higher weight to the first term in 
our loss function. Unless otherwise stated, in all the experiments we use two classifiers in our model.

To evaluate the models in a semi-supervised setting we adopt the standard practice in the literature \cite{semi-em}, thus, we use the stratified random sampling to sample a small set from the original training data to be used as the training set for the models. We repeat all the experiments 3 times with different random seeds, and report the average of the results.

\noindent\textbf{Evaluation metrics.} \AMAZONDT and \YELPDT datasets are balanced benchmarks, we report accuracy in these datasets. \AGNEWSDT dataset is a multi-class task, following \citet{do-pretrain} we report macro-F1 in this dataset. \ADRDT and \PRODUCTDT datasets are imbalanced. Following the argument made by \citet{perf-metric} about imbalanced datasets, we report the F1 measure in the minority (the positive) class to account for both the quality and the coverage of the models.


\section{Results and Analysis} \label{sec:result}

\subsection{Main Results}\label{subsec:main-results}
Table \ref{tbl:result-semi} reports the results of \METHOD and the baselines in all the datasets. We observe that \METHOD in all the cases is either the best or on a par with the best model. We particularly see that the improvement is substantial in \ADRDT dataset. This is, in part, due to the skewed class distributions in this dataset. Our model efficiently utilizes the entire set of unlabeled documents resulting in a higher recall, and at the same time, maintaining a high precision. We also inspected the documents in ADR task and observed that they are significantly more diverse than the ones in the other four tasks. This quality of ADR makes it specifically susceptible to the number of training examples. We also note that \textit{Mutual-learn} completely fails to learn in this dataset. Our investigations revealed that the extreme class imbalance is the underlying reason.\footnote{We subsampled from the positive set in \PRODUCTDT dataset and constructed a highly imbalanced dataset, this model yielded the same results in this case too.} 

\begingroup 
\setlength{\tabcolsep}{3pt} 
\begin{table}
\centering
\footnotesize
\begin{tabu}{p{0.1in} p{0.25in} p{0.2in} p{0.2in} p{0.2in} p{0.2in} p{0.2in} } \Xhline{2\arrayrulewidth}
 \cline{1-7} & & 
 \multicolumn{1}{c}{\textbf{Amaz.}} &
 \multicolumn{1}{c}{\textbf{\YELPDT~~~}} &
 \multicolumn{1}{c}{\textbf{AG-N.}} &
 \multicolumn{1}{c}{\textbf{\ADRDT~~}}  &
 \multicolumn{1}{c}{\textbf{Prod.}} \\  
 \cmidrule[\heavyrulewidth](l){3-7} 
 \multicolumn{1}{c}{\textbf{\# Doc}} & 
\multicolumn{1}{c}{\textbf{Model}} & 
\textbf{Acc} & \textbf{Acc} & \textbf{F1} & \textbf{F1} & \textbf{F1} \\ \Xhline{3\arrayrulewidth}
\multicolumn{1}{c}{\multirow{10}{*}{300}} & \multicolumn{1}{c}{\textit{Baseline}} & 0.815 & 0.891 & 0.863 & 0.238 & 0.728 \\ 
\multicolumn{1}{c}{} & \multicolumn{1}{c}{\textit{Self-train}} & 0.833 & 0.883 & 0.871 & 0.303 & 0.731 \\ 
\multicolumn{1}{c}{} & \multicolumn{1}{c}{\textit{Tri-train+}} & 0.867 & 0.914 & 0.873 & 0.306 & 0.734 \\ 
\multicolumn{1}{c}{} & \multicolumn{1}{c}{\textit{Mut-learn}} & 0.851 & 0.908 & 0.877 & 0.024 & 0.753 \\ 
\multicolumn{1}{c}{} & \multicolumn{1}{c}{\textit{Space-rep}} & 0.860 & 0.899 & 0.872 & 0.258 & 0.727 \\
\multicolumn{1}{c}{} & \multicolumn{1}{c}{\textit{Co-Deco.}} & - & - & - & 0.310 & 0.754 \\
\multicolumn{1}{c}{} & \multicolumn{1}{c}{\textit{HAU}} & 0.867 & 0.912 & 0.873 & 0.309 & 0.753 \\
\multicolumn{1}{c}{} & \multicolumn{1}{c}{\textit{UPS}} & 0.870 & 0.910 & 0.877 & 0.323 & 0.755 \\
\multicolumn{1}{c}{} & \multicolumn{1}{c}{\textit{BDD}} & 0.845 & 0.892 & 0.876 & 0.291 & 0.734 \\ 
\multicolumn{1}{c}{} & \multicolumn{1}{c}{\textit{Sel-Reg}} & 0.867 & 0.912 & 0.886 & 0.116 & 0.750 \\ 
\cmidrule(l){2-7}
\multicolumn{1}{c}{} & \multicolumn{1}{c}{\textit{\METHOD}} & \textbf{0.881} & \textbf{0.926} & \textbf{0.888} & \textbf{0.386} & \textbf{0.767} \\ \Xhline{3\arrayrulewidth}

\multicolumn{1}{c}{\multirow{10}{*}{500}} & \multicolumn{1}{c}{\textit{Baseline}} & 0.859 & 0.917 & 0.883 & 0.312 & 0.740 \\ 
\multicolumn{1}{c}{} & \multicolumn{1}{c}{\textit{Self-train}} & 0.865 & 0.916 & 0.885 & 0.335 & 0.741 \\ 
\multicolumn{1}{c}{} & \multicolumn{1}{c}{\textit{Tri-train+}} & 0.880 & 0.923 & 0.888 & 0.365 & 0.758 \\ 
\multicolumn{1}{c}{} & \multicolumn{1}{c}{\textit{Mut-learn}} & 0.880 & 0.920 & 0.889 & 0.108 & 0.767 \\ 
\multicolumn{1}{c}{} & \multicolumn{1}{c}{\textit{Space-rep}} & 0.862 & 0.917 & 0.888 & 0.295 & 0.737 \\
\multicolumn{1}{c}{} & \multicolumn{1}{c}{\textit{Co-Deco.}} & - & - & - & 0.345 & 0.766 \\  
\multicolumn{1}{c}{} & \multicolumn{1}{c}{\textit{HAU}} & 0.879 & 0.917 & 0.882 & 0.349 & 0.767 \\
\multicolumn{1}{c}{} & \multicolumn{1}{c}{\textit{UPS}} & 0.878 & 0.918 & 0.888 & 0.334 & 0.771 \\
\multicolumn{1}{c}{} & \multicolumn{1}{c}{\textit{BDD}} & 0.859 & 0.891 & 0.878 & 0.312 & 0.741 \\ 
\multicolumn{1}{c}{} & \multicolumn{1}{c}{\textit{Sel-Reg}} & 0.876 & 0.912 & \textbf{0.892} & 0.178 & 0.770 \\ 
\cmidrule(l){2-7}
\multicolumn{1}{c}{} & \multicolumn{1}{c}{\textit{\METHOD}} & \textbf{0.891} & \textbf{0.928} & 0.891 & \textbf{0.421} & \textbf{0.783} \\ \Xhline{3\arrayrulewidth}
\end{tabu}
\caption{Performance of \METHOD compared to the baselines in all the datasets. We follow previous studies and report Accuracy in \AMAZONDT and \YELPDT datasets; and report F1 in \AGNEWSDT\!, \ADRDT\!, and \PRODUCTDT datasets. The models were trained on 300 and 500 labeled documents. } \label{tbl:result-semi}
\end{table}
\endgroup 



\subsection{Empirical Analysis}\label{subsec:analysis}
In this section, we contrast \METHOD with domain specific language model pretraining, analyze the resistance of it to semantic drift, report an ablation study on the efficacy of the individual modules, examine the pretraining mechanism in \METHOD\!, analyze the hyper-parameter sensitivity, and analyze the convergence performance. 

We begin by validating our claim that our model can be complementary to language model pretraining (see Section \ref{sec:intro}). We compare \METHOD to domain specific language model pretraining \cite{do-pretrain}. Thus, we use the unlabeled data described in Section \ref{sec:setup} to pretrain \textit{Baseline} model using the masked language model and the next sentence prediction tasks \cite{bert}. Table \ref{tbl:result-lm} reports the results of this experiment. We observe that the combination of \METHOD and pretraining yields an additional improvement. This experiment and the next ones require running models for multiple times. We carried them out in the \ADRDT dataset with 500 initial labeled documents.

\begin{table}
\centering
\footnotesize
\begin{tabu}{p{1.2in}  p{0.35in} p{0.45in} p{0.35in} } \Xhline{3\arrayrulewidth}
\multicolumn{1}{c}{\textbf{Model}} & \textbf{F1} & \textbf{Precision} & \textbf{Recall} \\ \Xhline{3\arrayrulewidth}
\multicolumn{1}{c}{\textit{DS-pretraining}} & 0.352 & 0.330 & 0.421 \\ 
\multicolumn{1}{c}{\textit{\METHOD}} & 0.421 & 0.344 & 0.548 \\ 
\multicolumn{1}{c}{\textit{Combined}} & 0.443 & 0.397 & 0.507 \\ \Xhline{3\arrayrulewidth}
\end{tabu}
\caption{Results of domain specific language model pretraining (\textit{DS-pretraining}), \METHOD\!, and their combination.} \label{tbl:result-lm}
\end{table}


\begin{figure}
    \centering
    \begin{subfigure}[t]{0.50\linewidth}
        \centering
        \includegraphics[width=\linewidth]{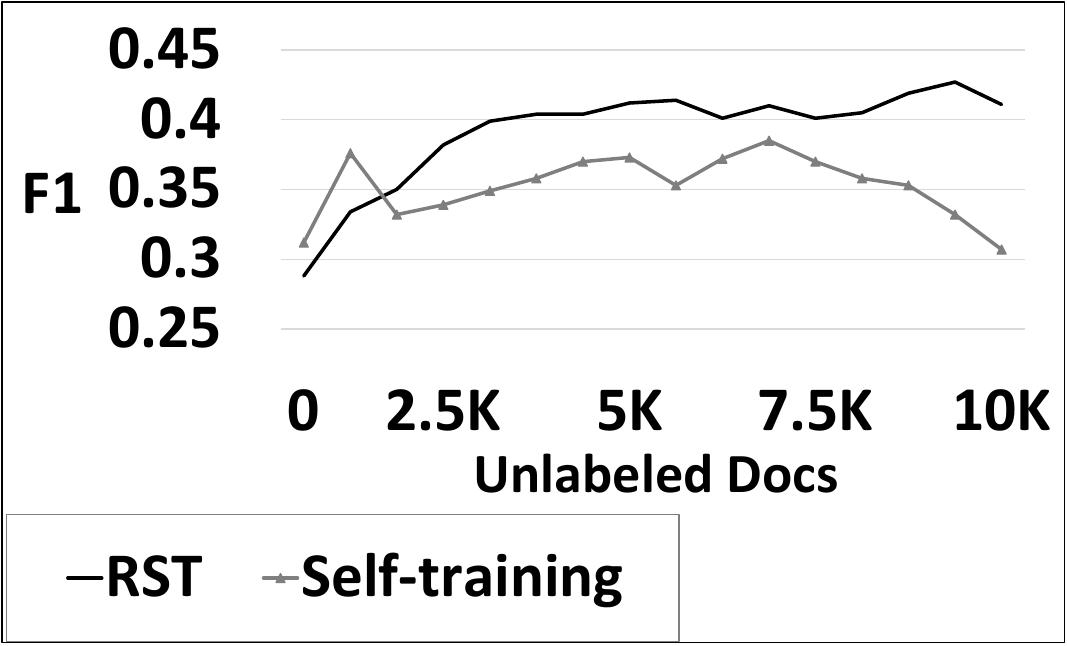}
        \caption{\METHOD vs self-training}
        \label{fig:curve-unlabeled}
    \end{subfigure}~~~
    \begin{subfigure}[t]{0.40\linewidth}
        \centering
        \includegraphics[width=\linewidth]{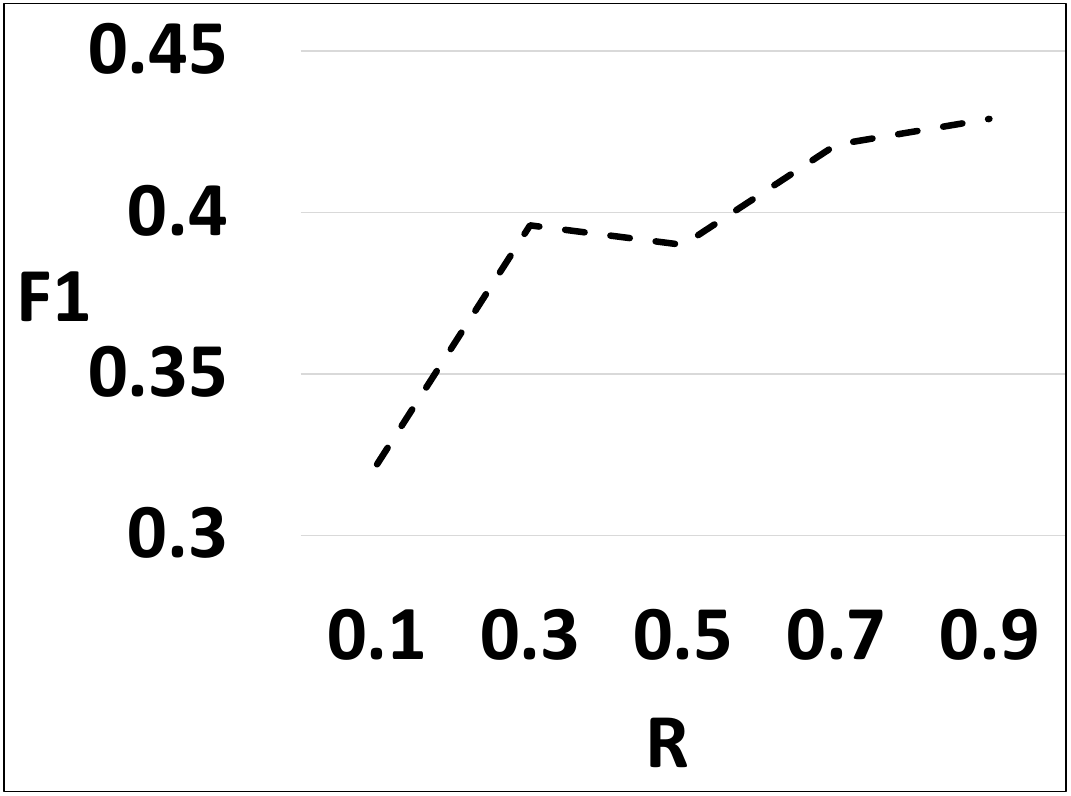}
        \caption{F1 vs sample-ratio}
        \label{fig:subsample-ratio}
    \end{subfigure}
    \caption{\textbf{\ref{fig:curve-unlabeled})} F1 of \METHOD and \textit{Self-pretraining} at varying unlabeled set sizes. \textbf{\ref{fig:subsample-ratio})} The sensitivity of \METHOD to the sample ratio. }\label{fig:h1-and-h2}
\end{figure}

To demonstrate the robustness of \METHOD against semantic drift we report the performance of \METHOD at varying number of added unlabeled documents during the bootstrapping iterations. The results are shown in Figure \ref{fig:curve-unlabeled}. We observe that in this regard our model is more robust compared to \textit{Self-training} baseline. We also see that our model reaches a plateau at about 3,500 unlabeled documents. Given that 10K unlabeled documents, used in our experiments, is a relatively large set for unsupervised text classification experiments \cite{neural-self-train-baselines}, this demonstrates that \METHOD is also data efficient.\footnote{The run-time of our model with 10,000 unlabeled documents was less than 3 hours using NVIDIA Titan RTX GPUs. We implemented the sequential variation of our model.}

Next, we report an ablation study on the efficacy of subsampling and pretraining steps. To do so, we replace subsampling with the regular confidence thresholding, and in another experiment, replace pretraining with the regular data augmentation. Table \ref{tbl:ablation} reports the results. We see that both strategies are effective, although pretraining makes a greater contribution. A fundamental question to answer is whether the effect of pretraining can be achieved by assigning a lower weight to pseudo-labels and augmenting them with labeled data. Table \ref{tbl:ablation-2} reports the results of this experiment when we replace pretraining with \textit{weighted augmentation} in \METHOD--we assigned the weight of 0.5 to the pseudo-labels.\footnote{Lower or higher weights does not yield an improvement.} We see that the performance substantially deteriorates, signifying the efficacy of pretraining strategy.

\begin{table}
\centering
\footnotesize
\begin{tabu}{p{1.0in}  p{0.35in} p{0.45in} p{0.35in} } \Xhline{3\arrayrulewidth}
\multicolumn{1}{c}{\textbf{Model}} & \textbf{F1} & \textbf{Precision} & \textbf{Recall} \\ \Xhline{3\arrayrulewidth}
\multicolumn{1}{c}{\textit{\METHOD}} & 0.421 & 0.344 & 0.548 \\ 
\multicolumn{1}{c}{\textit{\METHOD} w/o subsampling} & 0.394 & 0.289 & 0.624 \\ 
\multicolumn{1}{c}{\textit{\METHOD} w/o pretraining} & 0.357 & 0.292 & 0.498 \\ \Xhline{3\arrayrulewidth}
\end{tabu}
\caption{Ablation study on the efficacy of subsampling and pretraining techniques.} \label{tbl:ablation}
\end{table}

\begin{table}
\centering
\footnotesize
\begin{tabu}{p{1.0in}  p{0.35in} p{0.45in} p{0.35in} } \Xhline{3\arrayrulewidth}
\multicolumn{1}{c}{\textbf{Model}} & \textbf{F1} & \textbf{Precision} & \textbf{Recall} \\ \Xhline{3\arrayrulewidth}
\multicolumn{1}{c}{\textit{\METHOD}} & 0.421 & 0.344 & 0.548 \\ 
\multicolumn{1}{c}{Weighted augmentation} & 0.365 & 0.320 & 0.470 \\ \Xhline{3\arrayrulewidth}
\end{tabu}
\caption{F1, Precision, and Recall of \METHOD when pretraining is replaced with weighted data augmentation.} \label{tbl:ablation-2}
\end{table}

\begin{figure}
    \centering
    \begin{subfigure}[t]{0.40\linewidth}
        \centering
        \includegraphics[width=\linewidth]{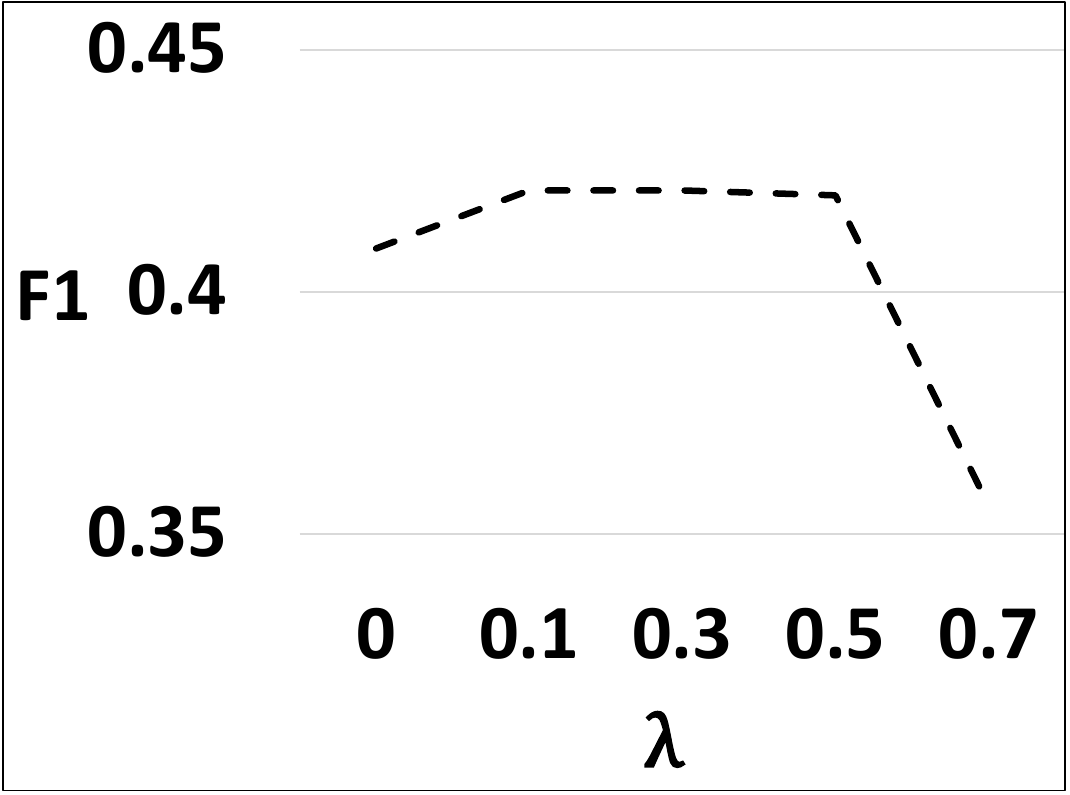}
        \caption{F1 vs $\lambda$}
        \label{fig:lambda}
    \end{subfigure}~~~
    \begin{subfigure}[t]{0.50\linewidth}
        \centering
        \includegraphics[width=\linewidth]{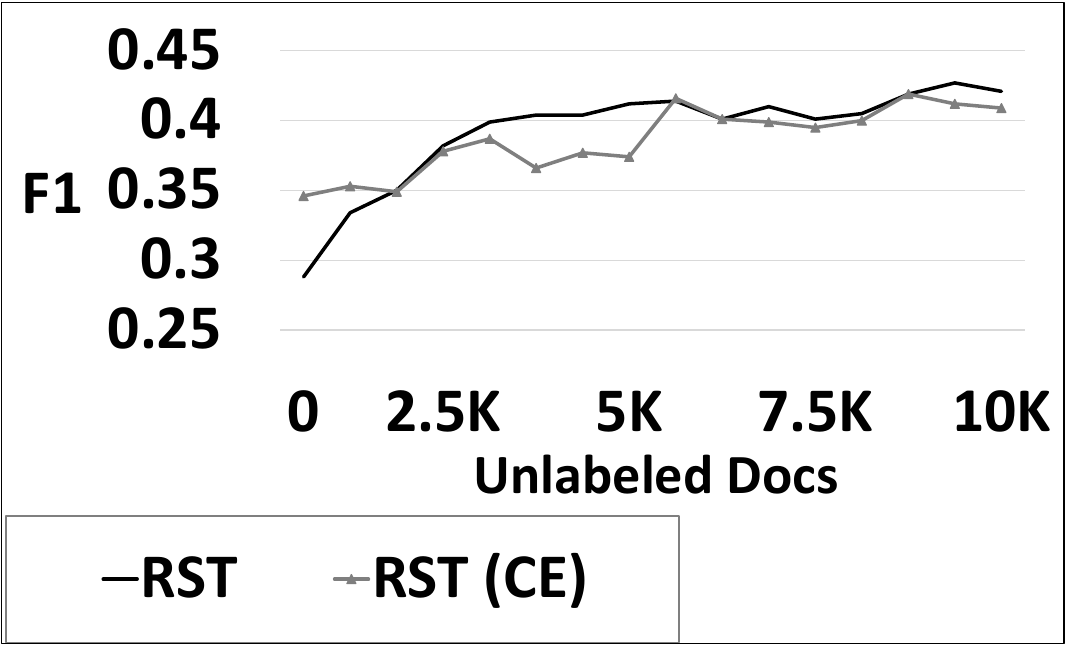}
        \caption{\METHOD vs \METHOD (CE)}
        \label{fig:convergence}
    \end{subfigure}
    \caption{\textbf{\ref{fig:lambda})}~The sensitivity of \METHOD to the penalty term $\lambda$. \textbf{\ref{fig:convergence})}~The convergence rate of \METHOD when we use the regular cross entropy instead of our loss function. The modified method is denoted by \METHOD (CE).}\label{fig:h1-and-h2}
\end{figure}

We now focus on hyper-parameter sensitivity. Figure \ref{fig:subsample-ratio} reports the sensitivity of our model to the sampling ratio in the subsampling stage. We see that after a certain threshold the performance reaches a plateau and the increase in performance is negligible. Figure \ref{fig:lambda} reports the performance of \METHOD at varying values of $\lambda$ in the objective function--Equation \ref{eq:loss}. This coefficient governs the impact of pseudo-labels. We see that as the value of $\lambda$ decreases, and a higher weight is assigned to the first term, the performance improves and ultimately drops again. This signifies the efficacy of our loss function, and verifies our argument in Section \ref{sub-sec:pretrain}. 

\begin{figure}
\centering
\includegraphics[width=0.50\linewidth]{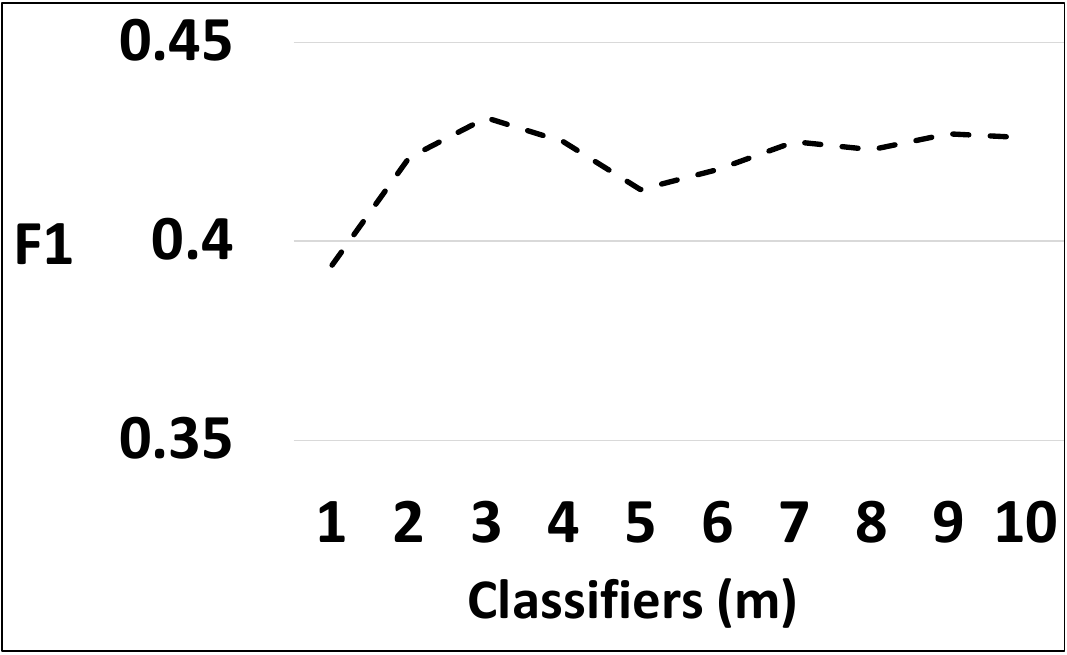}
\caption{The sensitivity of \METHOD to the number of classifiers. We see that our model reaches the highest performance when three classifiers are used.} \label{fig:cls-count}
\end{figure}

As we stated earlier, in all the experiments we used two classifiers in our model. To demonstrate the sensitivity of our model to the number of classifiers, we report the performance of \METHOD with varying number of classifiers. Figure \ref{fig:cls-count} illustrates the results. We see that by adding one more classifier our model can achieve slightly better results, however, after this cut-off the performance doesn't further improve. 

Our loss function (Equation \ref{eq:loss}) has two terms. The second term in the loss function ties the current training stage (using labeled data) to the training in the previous stage (using pseudo-labels). This raises the question whether this dependency makes the convergence speed slower. To answer this question, we replaced the entire objective with the regular cross entropy on labeled data. Figure \ref{fig:convergence} reports the results. We see that in terms of convergence, \METHOD is faster and more stable. This is perhaps due to catastrophic forgetting. Training on labeled data interferes with the already stored knowledge in the network and results in the fluctuations that we see in the new learning curve.

\begin{table}
\centering
\footnotesize
\begin{tabu}{p{0.9in}  p{0.25in} p{0.4in} p{0.3in} } \Xhline{3\arrayrulewidth}
\multicolumn{1}{c}{\textbf{Model}} & \textbf{F1} & \textbf{Precision} & \textbf{Recall} \\ \Xhline{3\arrayrulewidth}
\multicolumn{1}{c}{\textit{\METHOD}} & 0.421 & 0.344 & 0.548 \\ 
\multicolumn{1}{c}{Tri-training with entropy} & 0.351 & 0.324 & 0.383 \\ \Xhline{3\arrayrulewidth}
\end{tabu}
\caption{F1, Precision, and Recall of \METHOD compared to Tri-training. The Tri-training selection criterion is to select the pseudo-labels that have the least entropy.} \label{tbl:tritrain}
\end{table}



Table \ref{tbl:result-semi} compares our model with multiple baselines including several ensemble models, e.g., \textit{Tri-train+}, \textit{Mut-learn}, \textit{Co-Deco.}, and \textit{HAU}. As a reference point, one may still like to see how our model compares with an ensemble model armed with an entropy selection metric. Table \ref{tbl:tritrain} reports the results of this experiment. We see that \METHOD outperforms such a model, verifying our claims.

In summary, we evaluated our model in five standard datasets under two settings and compared with ten strong baselines. We showed that in all the cases our model is either the best or on a par with the best model. 
We plan to investigate the applicability of our model in cross-lingual settings.



\section{Conclusions} \label{sec:conclusion}

In this paper we proposed a semi-supervised text classifier. Our model is based on the self-training paradigm and employs neural network properties to enhance the bootstrapping procedure. Specifically, we use a subsampling technique to overcome the poor calibration of neural networks and to improve the candidate selection. Then, we exploit the catastrophic forgetting phenomenon in neural networks to alleviate the semantic drift problem. We evaluated our model in five public datasets and showed that it outperforms ten baselines.

\section*{Limitations} \label{sec:limit}


Our model is evaluated in standard English datasets for classification. As we stated earlier we plan to investigate the cross lingual setting in the next step.

The iterative nature of self-training imposes a high cost on the experiments. This has led to a few common practices. Most existing studies (including all the studies that we used as baselines) employ one underlying classifier to carry out the experiments--i.e., BERT or RNNs. This practice albeit limiting, is justified by the argument that if an algorithm does not make any assumption about the underlying structure of the classifier, then one can safely select the best available classifier and use it in the experiments. We used BERT in our experiments.

Another limitation is that, which is again stemmed from the high cost of self-training, one is typically forced to select a few sample sizes as labeled sets to carry out the experiments--e.g., 100 or 300. This is in contrast to similar research areas, such as Active Learning, when one can usually afford to report a learning curve to illustrate the performance with a few training examples all the way to using the full labeled dataset. Given that we have 10 baselines, we reported the performance with 300 and 500 labeled examples.





\bibliography{anthology,custom}
\bibliographystyle{acl_natbib}


\end{document}